\pdfoutput=1
\documentclass[11pt]{article}

\usepackage[final]{acl}

\usepackage{times}
\usepackage{latexsym}
\usepackage{amsmath}
\usepackage{adjustbox}
\usepackage{multirow}
\usepackage[T1]{fontenc}
\usepackage{listings}% http://ctan.org/pkg/listings
\usepackage[utf8]{inputenc}

\usepackage{microtype}

\usepackage{inconsolata}

\usepackage{graphicx}

\title{Spec-TOD: A Specialized Instruction-Tuned LLM Framework for Efficient Task-Oriented Dialogue Systems}

\author{
  Quang-Vinh Nguyen\thanks{Equal contribution.}, 
  Quang-Chieu Nguyen\footnotemark[1], 
  Hoang Pham,
  Khac-Hoai Nam Bui\thanks{Corresponding author.} \\
  Viettel Artificial Intelligence and Data Services Center, \\
  Viettel Group, Vietnam \\
  \{vinhnq29, chieunq, hoangpv4, nambkh\}@viettel.com.vn
}

\begin{document}
\maketitle
\begin{abstract}
Task-oriented dialogue (TOD) systems facilitate goal-driven interactions between users and machines. While recent advances in deep learning have improved the performance, TOD systems often struggle in low-resource scenarios with limited labeled data. To address this challenge, we propose Spec-TOD, a novel framework designed to train an end-to-end TOD system with limited data. Spec-TOD introduces two main innovations: (i) a novel specialized end-to-end TOD framework that incorporates explicit task instructions for instruction-tuned large language models (LLMs), and (ii) an efficient training strategy that leverages lightweight, specialized LLMs to achieve strong performance with minimal supervision. Experiments on the MultiWOZ dataset, a widely used TOD benchmark, demonstrate that Spec-TOD achieves competitive results while significantly reducing the need for labeled data. These findings highlight the potential of the proposed framework in advancing efficient and effective TOD systems in low-resource settings. 
\end{abstract}
\section{Introduction}
\begin{figure}[!t]
    \centering
    \includegraphics[width=\columnwidth]{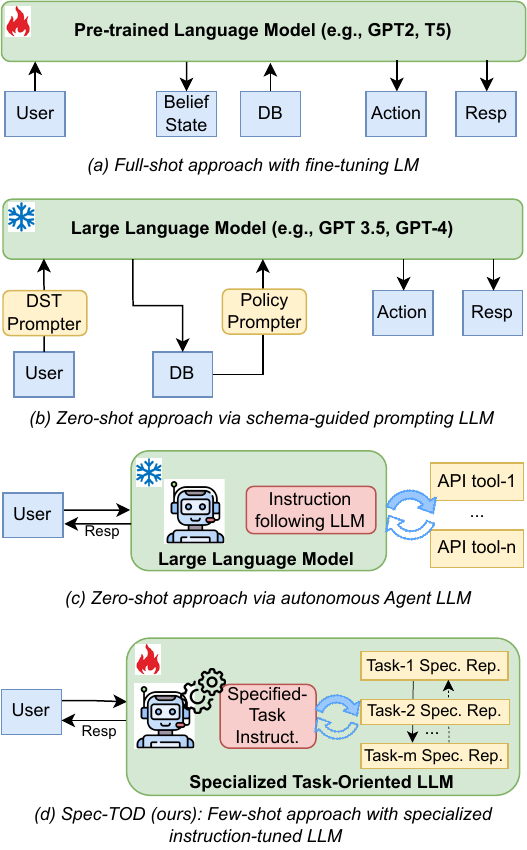} % Path to your image file
   \caption{Overview of end-to-end TOD Approaches: (a) traditional full-shot fine-tuning; (b) zero-shot LLM; (c) zero-shot Agent LLM; and (d) few-shot with instruction-tuned LLM (ours).}
    \label{fig:overview}
\end{figure}

Task-oriented dialogue (TOD) systems have emerged as a critical area of research in natural language processing (NLP), integrating multiple functionalities, such as understanding user input, dialogue state tracking, and response generation to effectively support users in accomplishing specific objectives \cite{HuangZG20}. Nevertheless, the development of these systems continues to pose significant challenges \cite{NiYPXC23}.
One of the primary obstacles in creating an effective TOD system is the requirement for extensive annotation and frequent retraining \cite{0001PCLY0CL23}. Specifically, prevailing TOD models either fine-tune pre-trained language models such as GPT-2 \cite{YangLQ21} and T5 \cite{Lee21,BangLK23} or prepare high-quality data curation for training pre-trained TOD language models \cite{HeDZWCLJYHSSL22,SuSMG0LZ22}, can achieve strong performance on TOD tasks (Figure \ref{fig:overview} (a)). 
The reliance on domain-specific annotated datasets poses challenges for reproducibility and scalability \cite{BaoWWSLMX23}. Despite progress in few-shot learning \cite{MiWL22, MoradshahiSL23}, current methods struggle to generalize across new domains and remain suboptimal. Recent advancements in large language models (LLMs), including proprietary models (e.g., GPT series \cite{abs-2303-08774}) and open-source models (e.g., Llama series \cite{abs-2302-13971}), have transformed NLP tasks by enabling zero-shot and few-shot generalization across various tasks, including the development of LLM-powered TOD systems (Figure \ref{fig:overview} (b)). Furthermore, emerging agent-based LLM technologies have reshaped multi-task learning in TOD systems by leveraging general-purpose, instruction-following models (e.g., GPT-4) that autonomously execute tasks through predefined external APIs \cite{XuMYSH24}. Nonetheless, current LLM-based approaches, including agentic LLMs, encounter two key limitations in end-to-end TOD systems: (i) The efficacy of zero-shot prompting, whether guided by schema or driven by API-based task specifications in agent-based architectures, is heavily contingent on the underlying LLM's capabilities. Competitive performance typically necessitates extremely large models, often with hundreds of billions of parameters. Such models entail significant computational costs, pose challenges for local deployment, and limit accessibility; (ii) Relying solely on instructional prompts to execute all TOD tasks, such as dialogue state tracking (DST) and natural language generation (NLG), may lack adaptability. This issue often results in suboptimal performance, especially in complex dialogue scenarios \cite{FengLLZW23}. Building on this analysis of current challenges, we formulate the following research question:

\textit{How can the performance of end-to-end TOD systems be improved by leveraging trainable LLMs with limited labeled data?}

To address this question, this study proposes Spec-TOD, a novel framework designed to optimize task-oriented dialogue systems by utilizing trainable LLMs for unifying multi-specified tasks via instruction tuning under constraints of limited labeled datasets. The main idea is to specify individual tasks with tailored instructions, enabling unified multi-task learning via instruction-tuned LLMs. Accordingly, Spec-TOD leverages multiple tasks of end-to-end TOD systems with specified instruction representations and a parameter-efficient tuning technique, enabling it to achieve competitive performance while utilizing much smaller models and limited computational resources. The primary contributions of this study are threefold: 
\begin{itemize}
    \item We introduce Spec-TOD, a novel framework tailored for TOD systems, leveraging trainable LLMs with task-specific instructions. To the best of our knowledge, Spec-TOD represents the first LLM-based framework explicitly designed for trainable end-to-end TOD systems.
    \item Spec-TOD employs parameter-efficient fine-tuning techniques with lightweight, open-source LLMs, enabling adaptability and cost-effective scalability for high-demand applications. The source code is available for further exploitation\footnote{https://github.com/quangvinh2110/Spec-TOD}. 
    \item We conduct extensive evaluations on multiple MultiWOZ versions, a widely used TOD benchmark. The results achieve promising performances compared with previous approaches, including full fine-tuning approaches with pre-trained language models and zero-shot approaches with large-scale LLMs. These findings highlight Spec-TOD’s potential to advance research in TOD systems.
\end{itemize}

\section{Literature Review}
\subsection{End-to-end task-oriented dialogue systems}
 
The traditional method follows a modular pipeline architecture, which is divided into four separate components: natural language understanding (NLU), dialogue state tracking (DST), dialogue policy learning (DPL), and natural language generation (NLG). However, this approach cannot leverage shared knowledge across all modules. Furthermore, the errors in earlier modules are propagated to later ones, leading to error  \cite{0001PCLY0CL23}. 
\begin{figure}[!ht]
    \centering
    \includegraphics[width=\columnwidth]{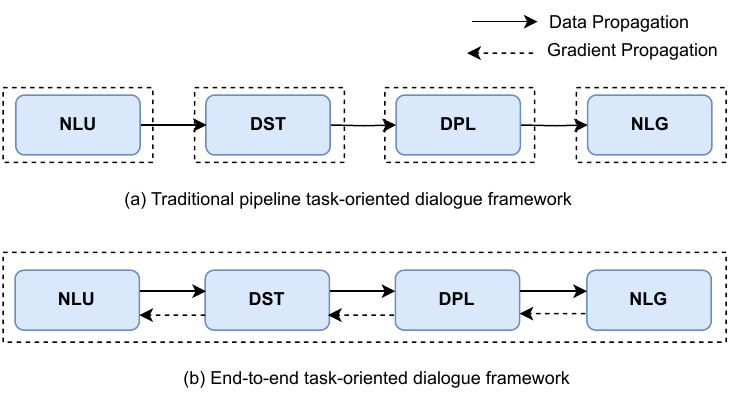} % Path to your image file
   \caption{Overview of TOD training methods. The solid line arrows and the dashed arrows represent data propagation and gradient propagation, respectively.}
    \label{fig:comparasion}
\end{figure}
To address this, the emerging models tend to shift to end-to-end TOD methods, as illustrated in the Figure \ref{fig:comparasion}. 
A key difference is that the end-to-end TOD methods can train all the components simultaneously by utilizing pre-trained language models (e.g., GPT-2 and T5) \cite{YangLQ21, YangHLE22, BangLK23, ChieuTB24}. Accordingly, those models are typically developed by fine-tuning the pre-trained language models to learn task-agnostic language representations on specific data.  

\subsection{LLM-powered zero-shot and few-shot TOD systems} 

Despite the promising results achieved through full fine-tuning of pre-trained language models, a significant challenge persists for current TOD systems: their development and maintenance demand substantial human annotation efforts for real-world deployment \citep{YangHLE22}. To address this issue, several studies have explored few-shot learning approaches to enhance the data efficiency of TOD systems \citep{LinMWF20, MiWL22, SuSMG0LZ22}. However, these methods often struggle to generalize effectively to unseen domains, limiting their broader applicability \citep{ImrattanatraiF23}.

Sequentially, the advent of LLMs has transformed this research landscape by leveraging instruction-following techniques. InstructTODS \citep{chung-etal-2023-instructtods}, one of the earliest LLM-powered TOD frameworks, employs in-context learning (ICL) to enable zero-shot, end-to-end TOD systems, offering improved flexibility and scalability for adapting to new domains. Subsequently, SGP-TOD \citep{ZhangPLZM23}, which operates on predefined task schemas (e.g., dialogue state tracking and natural language generation), introduces an instruction-following approach by utilizing zero-shot or few-shot prompting with several examples to enhance performance by providing comprehensive task-specific information. More recently, AutoTOD \citep{XuMYSH24}, an agent-based architecture, was proposed to autonomously execute tasks using instruction prompts and external predefined task APIs. However, the heavy reliance on underlying LLMs, which are often difficult to control, results in performance inferior to that of fully fine-tuned traditional approaches, particularly in complex scenarios \citep{HuaXJZSW023}.

To address these limitations, this study proposes a trainable, lightweight LLM framework tailored for end-to-end TOD systems. By integrating efficient training strategies and operating with limited labeled data, the proposed framework achieves competitive performance compared to existing approaches in this research field.

\section{Methodology}
This section establishes the foundation for our work by first providing background information on end-to-end TOD systems. Subsequently, we delve into the details of our proposed framework.
\subsection{End-to-end TOD formulation}
In traditional end-to-end TOD, at each turn $t$,  the DST module summarizes the dialogue state (Belief State) $B_t$ given the dialogue context $C_{t}$:
\begin{equation}
 C_{t}=\{(U_1, B_1,R_1),.., (U_{t-1}, B_{t-1},
 R_{t-1}), U_t\}
\end{equation} where $U_i$ and $R_i$ is user utterance and system response at turn $i$. The belief state $B_t$  is defined as the concatenation
of the domain/task (i.e., user intent) $d_t$ and a set of slot-value pairs
\begin{equation}
  B_t =  d_t \oplus \{(s_1,v_1),...,(s_n,v_n)\} 
\end{equation}
where $n$ denotes the total number of pairs in the set. The output belief states $B_t$ is then put into the NLG module for generating appropriate action $A_t$ and response $R_t$ based on database state $DB_t$:
\begin{equation}
\begin{aligned}
    DB_t = QUERY(B_t)
    \\
A_t, R_t = NLG(C_t, DB_{1}^{t},A_{1}^{t-1})
\end{aligned}
\end{equation}
Intuitively, the traditional formulation requires a substantial amount of labeled data across multiple tasks to ensure sufficient training coverage and achieve strong performance. To address this, we propose a new framework that specifies multiple tasks in the TOD system by reformulating each task with a dedicated instruction, enabling effective learning through few-shot fine-tuning.
\subsection{Spec-TOD Framework}
\begin{figure*}[ht]
    \centering
    \includegraphics[width=\textwidth]{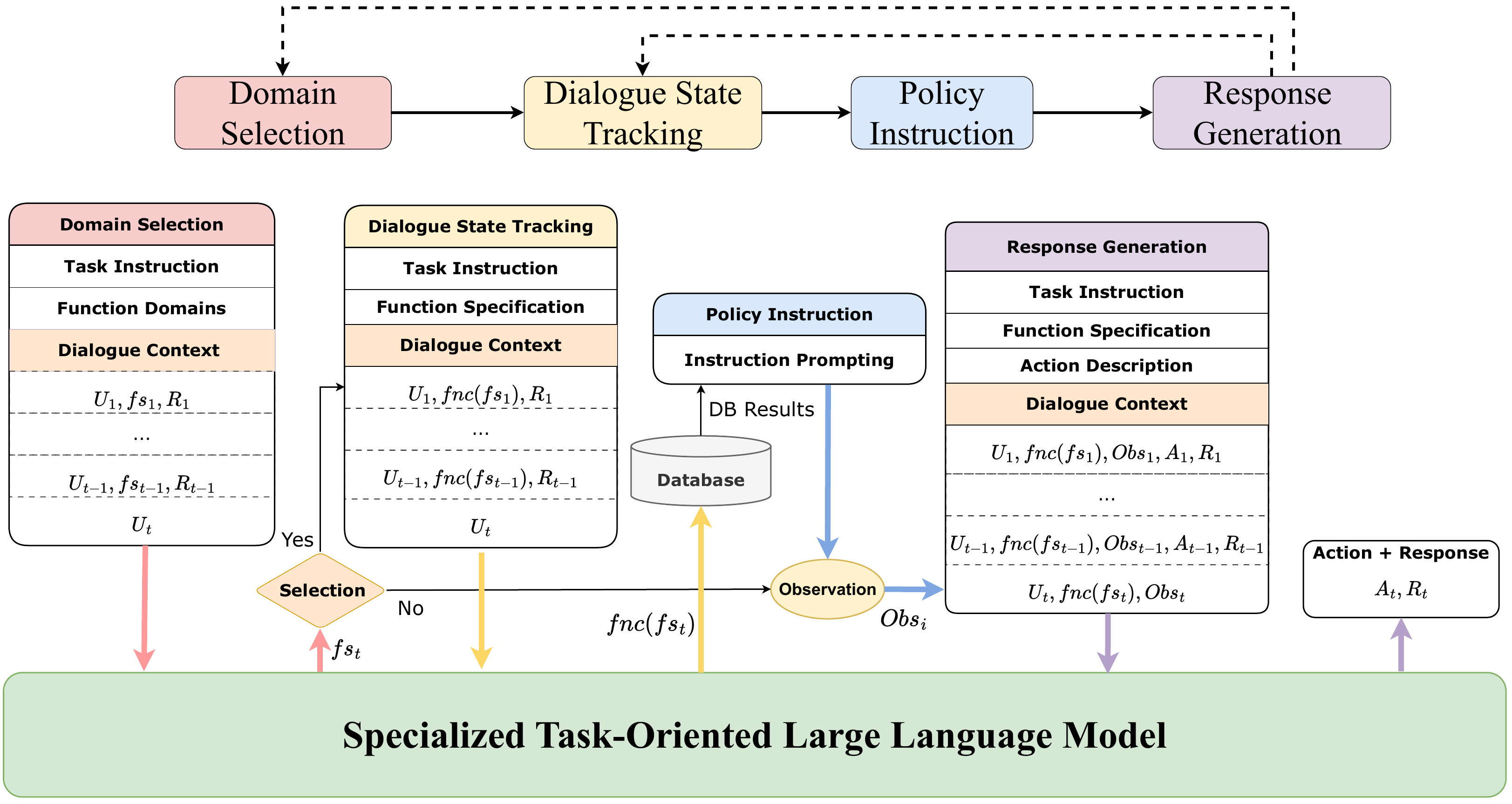} % Path to your image file
    \caption{Overview of the Spec-TOD framework, which includes four main specified tasks: Domain Selection, Dialogue State Tracking, Policy Instruction, and Generation Response. Accordingly, except for Policy Instruction is executed with predefined prompt instructions, other task functions are executed and updated with few-shot learning via instruction tuning LLMs.}
    \label{fig:architecture}
\end{figure*}

Figure \ref{fig:architecture} provides an overview of the proposed SpecTOD framework. The framework reformulates the four primary tasks of task-oriented dialogue (TOD) systems into a unified representation, enabling a cohesive multi-task learning approach through instruction-tuned LLMs.
The core idea is to conceptualize multi-task learning in end-to-end TOD systems as a function-calling paradigm. Training samples are reformatted to align with instruction-based templates, drawing inspiration from chat-tuned LLMs \cite{abs-2402-10466}. Accordingly, each dialogue domain is represented as a distinct function, with slot values within the domain serving as function arguments. Technically, SpecTOD integrates four core components: Domain Selection (DS), Dialogue State Tracking (DST), Policy Instruction (PI), and Response Generation (RG). These components are sequentially described as follows:

\subsubsection{Domain Selection}
Consider a set of functions representing distinct domains, denoted by $\mathcal{F} = \{f_0, f_1, \dots, f_n\}$. Each function $f_i \in \mathcal{F}$ corresponds to a specific domain (e.g., restaurant, taxi, hotel...) and is defined by three key attributes: a unique identifier, a brief domain description, and a set of arguments. Formally, a function $f_i$ is structured as follows:
\begin{quote}
\begin{verbatim}
f_i = {
"name": name of domain;
"description": domain description;
"arguments":{
"slot_name": name of slot;
"type": type of data;
"description": a textual explanation;
"possible_values": list of values}
}
\end{verbatim}
\end{quote}
The dialogue context for the domain selection task at turn $t$, denoted by $C_t(DS)$, encapsulates the sequence of interactions up to the current turn, which is re-formatted as follows:
\begin{equation}
    \begin{aligned}
        C_t(DS) = 
        \\
        \{(U_1,fs_{1},R_1),.., (U_{t-1},fs_{t-1},R_{t-1}),U_t\}
    \end{aligned}
\end{equation}
where $U_i$ represents the user utterance at turn $i$, $f_{s_i} \in \mathcal{F}$ denotes the selected domain at turn $i$, and $R_i$ signifies the system response at turn $i$. The term $U_t$ indicates the current user's utterance at turn $t$.
The domain selection task at turn $t$ involves determining the appropriate domain $f_{s_t}$ based on the dialogue context and task instructions. This process is modeled as:
\begin{equation}
\begin{aligned}
f_{s_t} = 
\\
\text{LLM} \left( \text{task\_inst\_DS} \oplus \mathcal{F} \oplus C_t(DS) \right)
\end{aligned}
\end{equation}
where $\text{task\_inst\_DS}$ refers to the instruction prompt for the domain selection task (detailed in the Appendix \ref{prompt-template}), $\oplus$ denotes the concatenation operation. The output $f_{s_t}$ is the selected domain at turn $t$, which serves as input for the subsequent dialogue state tracking task.
\subsubsection{Dialogue State Tracking}
The objective is to generate arguments for the domain function $f_{s_t}$  in JSON format. Specifically, the dialogue context for the DST task is reformatted as follows:
\begin{equation}
    \begin{aligned}
     C_t(DST) = \{(U_1,fnc(f_{s_1}),R_1)
 \\
 ,..,(U_{t-1},fnc(f_{s_{t-1}}),R_{t-1}),U_t\}
    \end{aligned}
\end{equation}
where $fnc(f_{s})$ represents the structured output of the DST task from previous turns, defined as:
\begin{quote}
\begin{lstlisting}
fnc($fs_i$)= {
"name": $fs_i$;
"argument":{($s_1$,$v_1$),..,($s_n$,$v_n$)}}
\end{lstlisting}
\end{quote}
where $(s_i,v_i)$ denotes slot-value pair. In this regard, the DST task is executed as follows:
\begin{equation}
\begin{aligned}
fnc(f_{s_{t}}) =
\\
LLM(task\_inst\_DST \oplus f_{s_t} \oplus C_t(DST)) 
\end{aligned}
\end{equation}
where $task\_inst\_DST$ refers to the instruction prompt for the DST task, as illustrated in the Appendix \ref{prompt-template}.

\subsubsection{Policy Instruction}
Following the execution of the DST task, defined as $fnc(f_{s_{t}})$, on the database, the system generates an observation denoted as $Obs_t$, which encapsulates the search results. This observation can manifest in two distinct forms. Firstly, $Obs_t$ may quantify the number of entities that fulfill the constraints specified by the function. Secondly, if the function remains unchanged between consecutive time steps (i.e., $fnc_{t-1} = fnc_t$) or if the function selected at time step $t$ is the null function (i.e., $f_{s_t} = f_0$), then $Obs_t$ may indicate that no function call is required, denoted as "Do not need to call function."

\subsubsection{Response Generation}
Based on the output of the DST task and the observation from the database, the system generates the action and response. The dialogue context of the task is formatted as follows:
\begin{equation}
\begin{aligned}
 C_t(GR)= 
    \{(U_1,fnc(f_{s_{1}}),Obs_1,A_1,R_1),
    \\
    .., (U_{t-1}, fnc(f_{s_{t-1}}), Obs_{t-1},A_{t-1},R_{t-1}), 
    \\ (U_t,fnc(f_{s_{t}}),Obs_t)\}
\end{aligned}
\end{equation}
where $A_i$ and $R_i$ denote the action and response of the system in the previous turn $i$, respectively. Sequentially, the task is executed as follows:
\begin{equation}
\begin{aligned}
    A_t, R_t = LLM( task\_inst\_GR \oplus Act_{des} 
    \\
    \oplus fnc(f_{s_t}) \oplus C_t(GR))
\end{aligned}
\end{equation}
where $task\_inst\_GR$ denotes the prompt instruction of the response generation task (illustrated in the Appendix \ref{prompt-template}). The response has to follow the predefined action description $Act_{des}$,  which is defined in each schema of the specific domain \cite{0001GR0ZW22}. In this study, based on the schema of the benchmark datasets \cite{BudzianowskiWTC18}, we specify the six descriptions into three typical actions, which are formulated as follows:
\begin{equation}
\begin{aligned}
Act_{des} =  <Info;Request;NoOffer;
\\
Recommend;Select;General>
\end{aligned}
\end{equation}
where:
\begin{itemize}
    \item \textbf{Info}:  provide information about an entity (if multiple matched results exist, choose one) in the form of [value\_xxx] if requested by the user.
    \item \textbf{Request}:  inform the number of available offers ([value\_choice]) and ask the user for more preferences on the requested entity to narrow down the search results.
    \item \textbf{NoOffer}: inform the user that no suitable offer could be found.
    \item \textbf{Recommend}: recommend an offer to the user and provide its information.
    \item \textbf{Select}: inform the user to select among options.
    \item \textbf{General}: greet and welcome the user, and inquire if there are anything further requirements.
\end{itemize}
\subsection{Instruction-Tuned LLM Strategy}
\begin{figure}[!ht]
    \centering
    \includegraphics[width=\columnwidth]{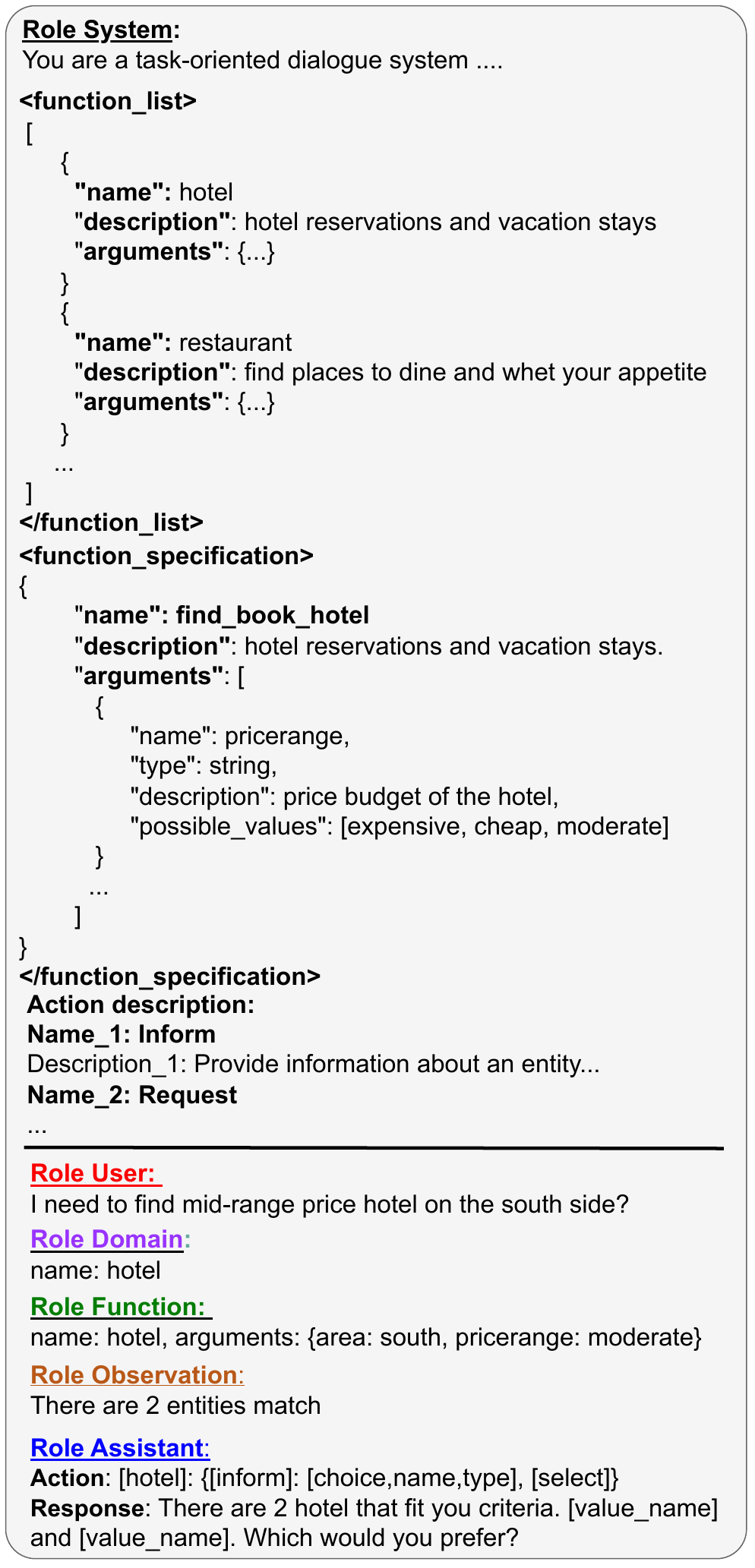} % Path to your image file
   \caption{An example of our prompt construction with 6 roles: System, User, Domain, Function, Observation, and  Assistant. The system prompt itself is comprised of four descriptions: (i) the end-to-end task, (ii) the list of function domains (for domain selection task), (iii) the selected function specification, and (iv) the action description.}
    \label{fig:format_sample}
\end{figure}

The primary goal of the proposed framework is to enhance the efficiency of multi-task learning in end-to-end TOD systems by fine-tuning an LLM with tailored instructions \cite{Xin0LZZWLH024}. To achieve this, we restructured dialogue samples into six distinct roles: system, user, domain, function, observation, and assistant. Rather than applying a uniform loss function across all tasks, we designated specific loss calculations to individual roles corresponding to each task. An illustrative example of the re-formatted dialogue sample, highlighting the six roles, is shown in Figure \ref{fig:format_sample}.
The loss function is computed by aggregating the losses derived from distinct tasks. Specifically, domain selection employs the "domain" role within the loss function framework. For dialogue state tracking, the loss function is defined for the "function" role, which is tasked with generating the function's arguments. In response generation tasks, the loss is associated with the "assistant" role, encompassing both action selection and response generation for the user. The overall loss function for the instruction tuning process is formulated as a composite of the aforementioned task-specific loss functions:
\begin{equation}
    \mathcal{L} = \mathcal{L}_{FS} + \mathcal{L}_{FC} + \mathcal{L}_{RG}
\end{equation}

\section{Experiments}
\subsection{Datasets and Implementation Setting}
For the benchmark dataset, we leverage various versions of the MultiWOZ dataset (2.0, 2.1, and 2.2), which contain multi-turn and multi-domain dialogues \cite{BudzianowskiWTC18, EricGPSAGKGKH20, zang-etal-2020-multiwoz, ye-etal-2022-multiwoz}. Each dataset version comprises 8438, 1000, and 1000 samples for training, development, and testing, respectively. The primary distinction between these versions lies in their dialogue annotation, with later versions addressing errors and inconsistencies present in earlier iterations.
In the configuration, we employ LLaMA-3-8B \cite{grattafiori2024llama3herdmodels}, as the lightweight backbone LLM model, and instruction-tuned with Low-Rank Adaptation (LoRA) using a small rank alpha. Our fine-tuning was restricted to the $q\_proj$ and $v\_proj$ modules. The fine-tuning process spanned four epochs, with a peak learning rate of $3 \times 10^{-4}$ and a global batch size of 8. The context length is limited to 4096 tokens.
%  \footnote{https://huggingface.co/meta-llama/Meta-Llama-3-8B-Instruct}
% \subsection{Metrics and Implementation Setting}
As for evaluation, we follow previous works by adopting a combination of \textbf{Inform}, \textbf{Success}, and \textbf{BLEU} scores. The inform rate assesses the system’s accuracy in providing relevant entities, while the
success rate measures its ability to fulfill all user-requested slots. The combined score can be formulated as follows: 
\begin{equation}
    Combined = BLEU + 0.5* (Inform + Success)
\end{equation}
Furthermore, the Joint Goal Accuracy (\textbf{JGA}) metric is used as the benchmark measurement for the DST task. 
% Here, a function is considered correct only when all key-value pairs precisely match the goal pairs.

\subsection{Main Results}
Table \ref{tab:results} presents the main results of our proposed framework compared to state-of-the-art TOD models under various training data settings. 
\begin{table*}[ht]
\centering
\begin{adjustbox}{max width=\textwidth}
\begin{tabular}{|l|c|c|c|c|c|c|c|c|}
\hline
\textbf{Model} & \multicolumn{4}{|c|}{\textbf{MultiWOZ 2.0}} & \multicolumn{4}{|c|}{\textbf{MultiWOZ 2.2}} \\
\hline
% \rowcolor{lightgray}
& \textbf{BLEU} & \textbf{Success} & \textbf{Inform} & \textbf{Combined} & \textbf{BLEU} & \textbf{Success} & \textbf{Inform} & \textbf{Combined} \\
\hline
\multicolumn{9}{|l|}{\textbf{Full-shot fine-tuning}} \\
\hline
SimpleTOD \cite{Hosseini-AslMWY20} & 15.0 & 70.1 & 84.4& 92.2&  - &  -   &  -   & -    \\
UBAR \cite{YangLQ21}& 17.0 & 80.7    & 95.4   & 105.1 & 17.6 & 70.3 & 83.4&  94.4  \\
GALAXY \cite{HeDZWCLJYHSSL22} &  20.8 & 84.9 & 93.5& 110.0&  19.6 & 75.7    & 85.4   & 100.2   \\
TOATOD \cite{Sun0W023}      & - & - & -& -&   17.0 & 79.8    & 90.0   & 101.9   \\ 
Mars \cite{Sun0W023}    &19.9  & 78.0    & 88.9   &  103.4  & 19.6 & 78.0& 88.9& 103.0   \\ 
% PPTOD \cite{SuSMG0LZ22}     & 18.6 & 79.4 & 89.2& 102.9&    - & -   & -   & -   \\ 
\hline
\multicolumn{9}{|l|}{\textbf{Zero-shot prompting with LLMs (e.g., Llama-2-70B, GPT-3.5 and GPT-4)}} \\
\hline
SGP-TOD-GPT-3.5 \cite{ZhangPLZM23} & 9.0 & 69.8 & 83.8& 85.9 & 9.2 & 72.5 & 82.0 & 86.4 \\
% AutoTOD-GPT-3.5 \cite{XuMYSH24} & 9.3 & 82.8 & 87.2 & 94.3 &  - & -  & -  & - \\
AutoTOD-GPT-4 \cite{XuMYSH24} & 10.4 & 84.4 & 91.7 & 98.5 & - & -  & -  &  -\\
AutoTOD-Llama-2-70B \cite{XuMYSH24} & 7.8 & 69.8 & 73.3 & 79.4 &  - & -  & -  & -  \\
% Spec-TOD-Llama-3-70B (ours) &  &  &  &  &  &  &  &  \\
\hline
\multicolumn{9}{|l|}{\textbf{Few-shot with pretrained TOD models}} \\
\hline
MinTL  \cite{LinMWF20} (10\%)& 15.6 & 55.5 & 44.9 & 65.8 &- &- &- &- \\
PPTOD \cite{SuSMG0LZ22} (10\%)& 15.7 & 53.7 & 68.3 & 76.7 & -&- &- &- \\
Mars-G \cite{Sun0W023}  (10\%)  &15.6  & 55.3    & 69.4   &  78.0  & - & - & - & -   \\ 
\hline
\multicolumn{9}{|l|}{\textbf{Few-shot with Spec-TOD Framework (Ours)}} \\
\hline
% LLama-3-8B & - & - & -& - & -& - & - & - \\
% Qwen-2.5-7B  & -& - & -&- & - & -& - & - \\
Spec-TOD-LLama-3-8B (1\%) & 7.41 & 63.2 & 73.0 & 75.5 & 6.6 & 59.5 & 75.7 & 74.2 \\
Spec-TOD-LLama-3-8B (5\%) & 9.1 & 63.9 & 79.2 & 80.6  & 9.2 & 66.4 & 80.5 & 82.6 \\
Spec-TOD-LLama-3-8B (10\%) & 10.4 & 75.5 & 86.0 & 91.2 & 10.4& 77.1 & 87.2 & 92.6 \\
% Spec-TOD-Qwen-2.5-7B (ours) & -& - & -&- & - & -& - & - \\
\hline
\end{tabular}
\end{adjustbox}
\caption{Performance comparison on few-shot End-to-end benchmark on MultiWOZ 2.0 and MultiWOZ 2.2 across various training samples of the datasets. The difference in mean is statistically significant (p < 0.01).}
\label{tab:results}
\end{table*}
We derive the following key insights based on these results:

i) \textbf{Spec-TOD consistently outperforms prior few-shot models}: The experiment on MultiWOZ 2.0 datasets for few-shot strategies indicates the promising results of the proposed method, particularly in terms of Success and Inform scores, which are critical for evaluating task completion and dialogue informativeness. Compared to models like MinTL, PPTOD, and Mars-G, Spec-TOD demonstrates a significant improvement. For example, using only 10\% of the training data, Spec-TOD-LLaMA-3-8B achieves 75.5\% Success and 86.0\% Inform on MultiWOZ 2.0, while the best of the few-shot baselines (Mars-G) only reaches 55.3\% Success and 69.4\% Inform. This indicates that Spec-TOD is more effective at learning how to complete user goals and convey correct information with limited data. 

ii) \textbf{Spec-TOD matches full-shot models in Success and Inform scores}: The proposed method achieves comparable or even better performance in some cases, despite using only 10\% of the data. For instance, on MultiWOZ 2.2, Spec-TOD reaches 77.1\%  Success and 87.2\%  Inform, which achieves competitive results with full-shot models such as GALAXY (75.7\% Success, 85.4\% Inform). This highlights the data efficiency of Spec-TOD, achieving strong generalization with minimal examples.

iii) \textbf{Spec-TOD surpasses zero-shot prompting with large-scale LLMs in domain-specific tasks:} Despite their scale and general reasoning capabilities, LLMs such as GPT-3.5, GPT-4, and LLaMA-2-70B (with the exception of AutoTOD with GPT-4) underperform in domain-specific TOD. These results indicate that prompting alone cannot fully capture the structured behavior needed for such tasks, whereas Spec-TOD’s lightweight adaptation more effectively grounds LLMs in domain-specific contexts.

iv) \textbf{Spec-TOD demonstrates strong generalization capabilities across datasets}: Despite the increased complexity and annotation consistency in MultiWOZ 2.2, Spec-TOD maintains or improves its performance when transitioning from MultiWOZ 2.0. With 10\% data, it improves from 75.5\%  to 77.1\% in Success and 86\% to 87.2\% in Inform from MultiWOZ 2.0 to 2.2. This robustness to data distribution shifts underscores the flexibility of the Spec-TOD framework.

v) \textbf{Spec-TOD remains highly effective in extremely low-resource settings}: Even using only 1\% of the training data, Spec-TOD achieves 63.2\% Success and 73.0\% Inform on MultiWOZ 2.0, performing comparably to some larger zero-shot LLMs such as AutoTOD-LLaMA-2-70B (Success 69.8\%, Inform 73.3\%). Performance improves steadily with more data (e.g., at 5\% data samples: 63.9\% Success, 79.2\% Inform), demonstrating graceful scalability and reinforcing its practicality for domains with limited annotation budgets.
\begin{table}[h]
\centering
\begin{tabular}{|l|c|c|}
\hline
\textbf{Criteria/Model} & \textbf{PPTOD} & \textbf{Spec-TOD} \\ \hline
% Interesting ($\uparrow$)            & 2.5                 & \textbf{3.29}       \\ \hline
% Engaging ($\uparrow$)                & 2.33                & \textbf{3.1}        \\ \hline
Understandable ($\uparrow$)         & 3.66                & 4.3        \\ \hline
Relevant ($\uparrow$)                & 3.29                & 4.1        \\ \hline
Correct ($\uparrow$)                 & 2.76                & 3.75       \\ \hline
Appropriate ($\uparrow$)             & 3.42                & 4.15       \\ \hline
Fluently ($\uparrow$)                & 3.54                & 4.18       \\ \hline
Direct ($\uparrow$)                  & 2.51                & 3.32       \\ \hline
% \textbf{BLEU}                   & \textbf{15.25}               & 10.43       \\ \hline
\end{tabular}
\caption{Evaluation of GPT-score ranging from 0 to 5 for PPTOD and Spec-TOD across various criteria, using 10\% of the training dataset. Detailed descriptions of each criterion are provided in the Appendix \ref{GPT-Score}.
}
\label{table:gpt_score}
\end{table}

Observationally, the BLEU score measures n-gram precision with a brevity penalty to ensure fluency and accuracy, which might affect the final results in case of the same semantic meaning between the predicted answer and the reference answer. Consequently, we execute another experiment by using GPT scores \cite{ZhengC00WZL0LXZ23}, a metric that uses a very large LLM (e.g., GPT-4o) 
serves as an evaluator to judge whether the predicted answer shares the same semantics as the reference answer, to evaluate the output of our proposed approach. 
The results of this evaluation are reported in Table \ref{table:gpt_score}.
Accordingly, although the BLEU score is lower (10.4 vs. 15.7), GPT-score reflects better semantic accuracy.

\subsection{Detailed Analysis}
\subsubsection{Impact of LLM Backbone}
To validate the impact of different backbones on our approach, we conducted experiments with instruction versions of two additional modern open-source LLMs with a parameter size of 8 billion parameters, such as Qwen2 \cite{abs-2407-10671} and LLama-3.1-8B\footnote{https://ai.meta.com/blog/meta-llama-3-1/}, as detailed in Table \ref{tab:comparison of different backbones}.
\begin{table}[!h]
\centering
\begin{adjustbox}{max width=\columnwidth}
\begin{tabular}{|l|c|c|c|c|c|}
\hline
\textbf{Backbone}         & \textbf{BLEU}  & \textbf{Success} & \textbf{Inform} & \textbf{Comb.} & \textbf{JGA}   \\ \hline
% \rowcolor[HTML]{C0C0C0}
\multicolumn{6}{|c|}{\textbf{MultiWOZ 2.1}} \\ \hline
Qwen2    & 15.53 & 58.5    & 76.3   & 82.93    & 43.39 \\ \hline
LLama3  & 10.41 & 75.5    & 86.0   & 91.18    & 47.06 \\ \hline
LLama3.1  & 8.60 & 76.9    & 85.2   & 89.65    & 44.4 \\ \hline
% \rowcolor[HTML]{C0C0C0}
\multicolumn{6}{|c|}{ \textbf{MultiWOZ 2.2}} \\ \hline
Qwen2    & \textbf{15.88} & 58.6    & 75.7   & 83.03    & 45.66 \\ \hline
LLama3  & 10.41 & 77.1    & 87.2   & 92.56    & 47.68 \\ \hline
LLama3.1  & 8.65 & 77.5    & 84.7   & 89.75    & 45.8 \\ \hline
\end{tabular}

\end{adjustbox}
\caption{Comparison of Qwen2, LLama3 and LLama3.1 backbone performance when training with 10\% on MultiWOZ 2.1 and MultiWOZ 2.2 datasets.}
\label{tab:comparison of different backbones}
\end{table}
Accordingly, despite using different backbone models, the proposed framework consistently achieves strong performance across both MultiWOZ 2.1 and 2.2 datasets. This demonstrates the robustness of our approach, maintaining high success, inform, and combined scores regardless of the underlying backbone.

\subsubsection{Impact of Lora Configuration}

We conduct experiments on various LoRA configuration settings, as detailed in Table \ref{tab:comparation lora setting}. Accordingly, the proposed method does not benefit from a high number of fine-tuned parameters. 
\begin{table}[!h]
\centering
\renewcommand{\arraystretch}{1.5}
\small
\setlength{\tabcolsep}{2pt}
\begin{tabular}{|c|c|c|c|c|}
\hline
% \rowcolor{headergray}
\textbf{Lora-rank} & \textbf{Lora-alpha} & \textbf{Combined} & \textbf{JGA}   & \textbf{Fn\_Se} \\ \hline
 16 & 16 & 86.37 & 46.85 & 94.02 \\ \hline
                          32 & 16 & \textbf{92.56} & 47.68 & \textbf{94.75} \\ \hline
                          64 & 32 & 90.26 & \textbf{47.88} & 94.42 \\ \hline
                          64 & 128 & 91.64 & 45.66 & 94.17 \\ \hline
\end{tabular}
\caption{Performance with varying LoRA-rank and LoRA-alpha settings, trained on 10\% of the MultiWOZ 2.2 dataset. $Fn_{Se}$ denotes the metric accuracy for the domain selection task.}
\label{tab:comparation lora setting}
\end{table}
We hypothesize that with lower LoRA rank and alpha settings, the fine-tuned parameters are insufficient for effectively compressing useful information from the training data. In contrast, high LoRA rank and alpha configurations may lead to a loss of generalization in our model. Accordingly, the most suitable configuration is a LoRA rank of 32 and a LoRA alpha of 16, balancing the trade-off between information compression and generalization.

\subsubsection{Exploiting Number of Training Samples}
We evaluated the performance of Spec-TOD using 10\%, 50\%, and 100\% of the training data, respectively, as shown in Table \ref{tab:comparation with others}. 
\begin{table}[!ht]
\centering
\begin{adjustbox}{max width=\columnwidth}
\begin{tabular}{|l|c|c|c|c|c|}
\hline
% \rowcolor{headergray}
\textbf{Samples}     & \textbf{BLEU}  & \textbf{Success} & \textbf{Inform} & \textbf{Combined} \\ \hline
10\%           & \textbf{10.41} & 77.1   & 87.2   & 92.56    \\ \hline

50\%              & 10.09 & \textbf{83.5}    & \textbf{90.5}   & \textbf{97.09}    \\ \hline

100\%            & 10.22 & 83.4    & 90.3   & 97.07    \\ \hline
\end{tabular}
\end{adjustbox}
\caption{Performance comparison with various numbers of training samples on MultiWOZ 2.2 dataset.}
\label{tab:comparation with others}
\end{table}
Notably, when fine-tuning with the full dataset (100\% samples), the performance remains consistent with 50\% of the training data. This finding demonstrates that our approach does not require extensive amounts of data to achieve performance comparable to previous full-training approaches. 
Accordingly, with only 10\% of the training data, our model matches the performance of prior approaches that utilized the entire dataset. These results highlight the efficiency and robustness of our model in low-data scenarios.

\subsubsection{DST Task Performance}
\begin{figure}[h]
    \centering
\includegraphics[width=\columnwidth]{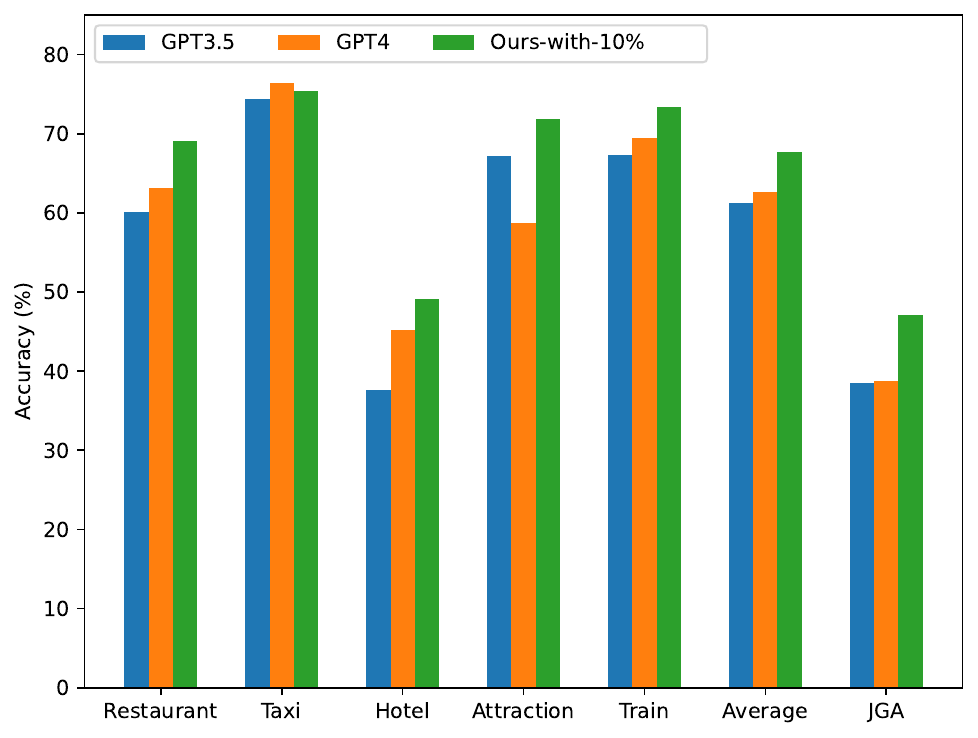}
    \caption{Performance comparison of our proposed method, training on 10 \% data with GPT-3.5 and GPT-4 on DST task across different domains.}
    \label{fig:fnc_cal}
\end{figure}

To further exploit the abilities of the proposed Spec-TOD, we evaluate the performance of the Function Calling (for DST) task, which is illustrated in Figure \ref{fig:fnc_cal}.
Accordingly, the reported results indicate that with training on  10\% of the data, Spec-TOD with Llama-3-8B outperforms GPT-3.5 and GPT-4 under zero-shot prompting.

\section{Conclusion}
TOD systems have long posed challenges in NLP due to limited flexibility and scalability to unseen domains. Recent emerging LLM technologies have revolutionized NLP applications by enabling zero-shot and few-shot generalization capabilities. However, utilizing LLMs for tasks requiring
knowledge grounding, such as TOD, continues to be a critical challenge that warrants further study, especially for end-to-end TOD systems. Therefore, this study presents a novel few-shot learning method for end-to-end TOD systems via instruction tuning with open-weight LLMs. The evaluation of various versions of the MultiWoz datasets indicates promising results of our proposed method with 10\% of the training samples.

% Bibliography entries for the entire Anthology, followed by custom entries
%\bibliography{anthology,custom}
% Custom bibliography entries only
\bibliography{custom}

\appendix

\section{Appendix}
\label{sec:appendix}
\subsection{GPT-score Criteria}
\label{GPT-Score}
Following the work in \cite{FuNJ024}, we define the GPT-Score with six criteria for the measurement as follows:
\begin{itemize}
    \item \textbf{Understand}: Are the responses understandable?
    \item \textbf{Relevant}:  Are the responses relevant to the conversation?
    \item \textbf{Correct}: Are the responses correct to conversations?
    \item \textbf{Appropriate}: Are the responses semantically appropriate?
    \item \textbf{Fluently}: Are the responses fluent?
    \item \textbf{Direct}: Are the responses directly answering the request from the input query?
\end{itemize}

\subsection{Specified Task Prompt Templates}
\label{prompt-template}
\begin{figure*}[!t]
\includegraphics[width=\textwidth]{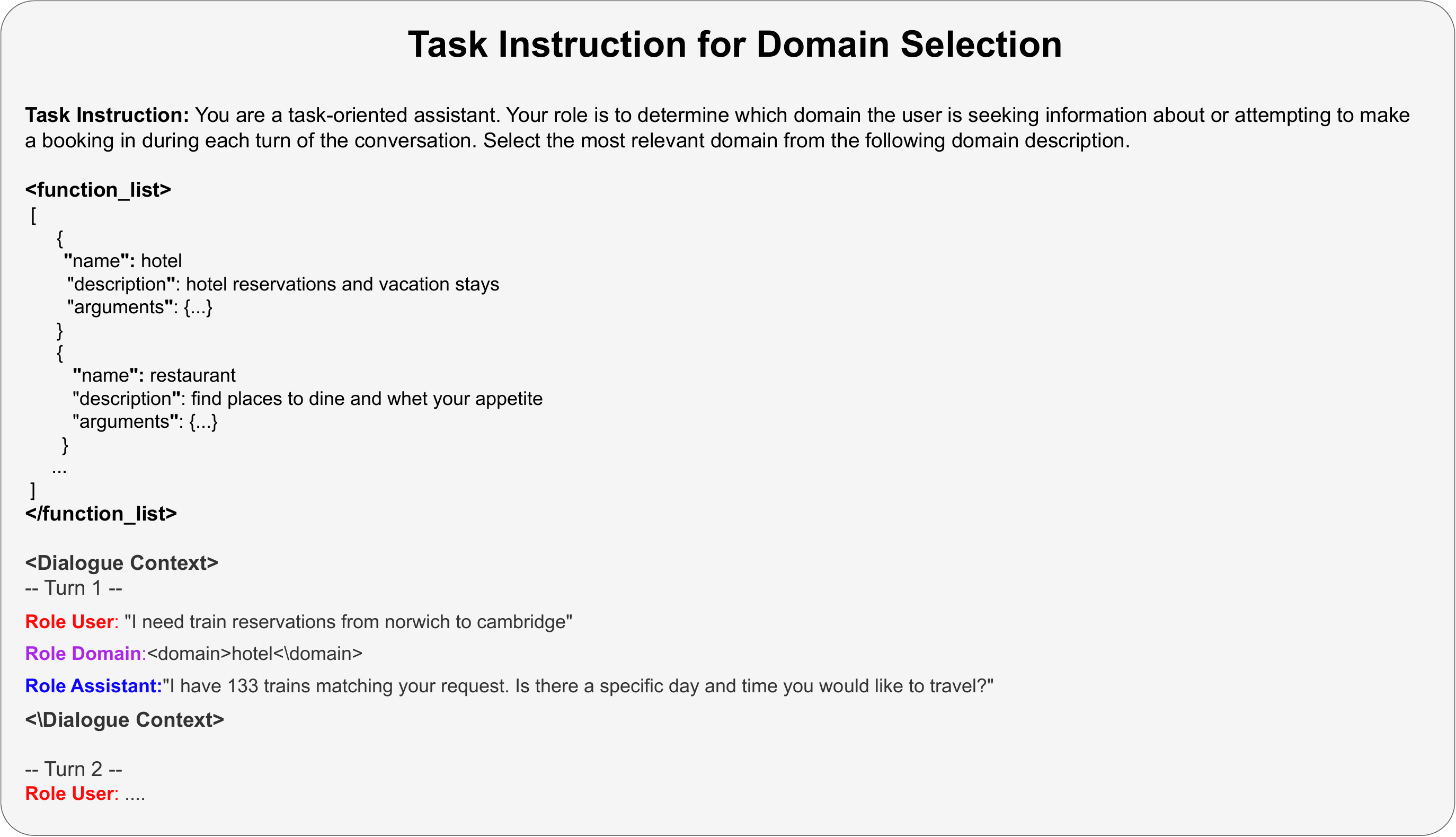}
\caption{Prompt template for Domain Selection task}
\label{fig:ds_prompt}
\end{figure*}

\begin{figure*}[!t]
\includegraphics[width=\textwidth]{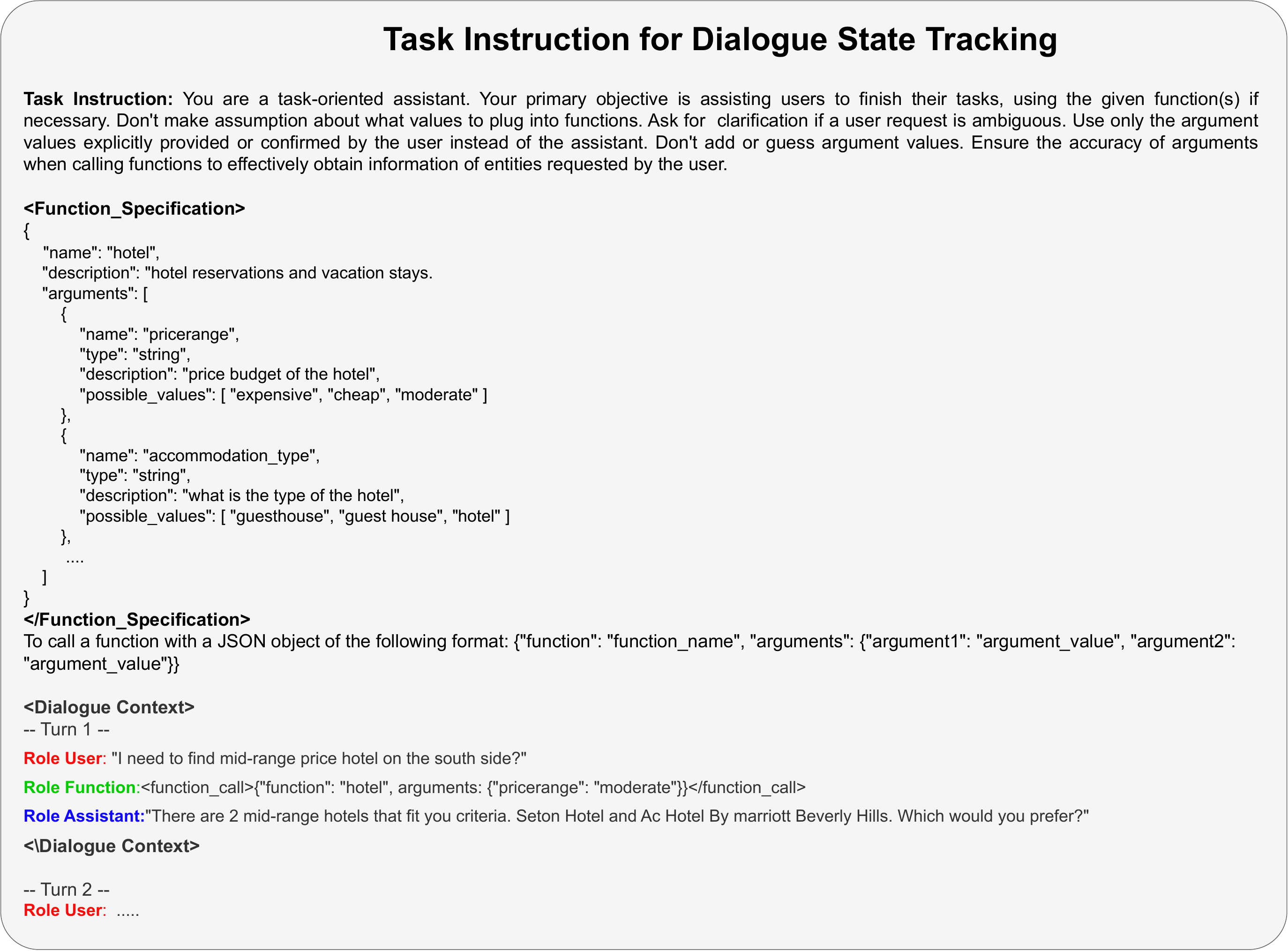}
\caption{Prompt template for Dialogue State Tracking task}
\label{fig:dst_prompt}
\end{figure*}

\begin{figure*}[!t]
\includegraphics[width=\textwidth]{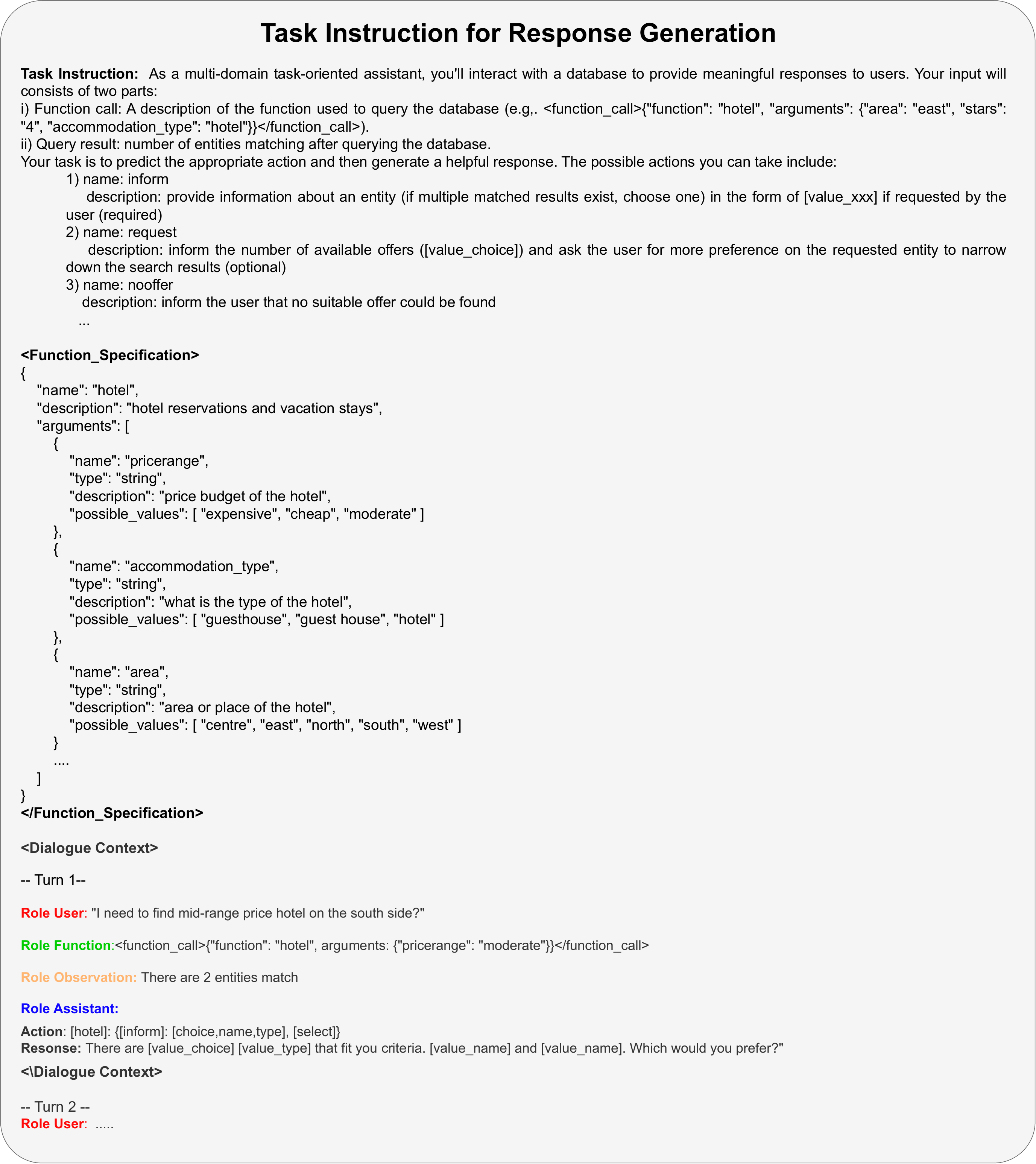}
\caption{Prompt template for Response Generation task}
\label{fig:rs_prompt}
\end{figure*}

For better reproducibility, we present all prompt templates in the appendix. Below is a quick reference list outlining the prompt templates and their usages:
\begin{itemize}
    \item Figure \ref{fig:ds_prompt}: Prompt the task instruction for Domain Selection Task.
    \item Figure \ref{fig:dst_prompt}: Prompt the task instruction for Domain Dialogue State Tracking Task.
    \item Figure \ref{fig:rs_prompt}: Prompt the task instruction for Response Generation Task.
\end{itemize}

\end{document}